\documentclass[10pt, a4paper]{article}

\usepackage[final]{lrec2026} 

\title{AgriChain: Visually-Grounded Expert-Verified Reasoning for Interpretable Agricultural Vision–Language Models}

\name{Hazza Mahmood, Yongqiang Yu, Rao Anwer}

\address{Mohamed bin Zayed University of Artificial Intelligence\\
Abu Dhabi, United Arab Emirates\\
{(hazza.mahmood, yongqiang.yu, rao.anwer)}@mbzuai.ac.ae}

\abstract{
Accurate and interpretable plant disease diagnosis remains a major challenge for vision–language models (VLMs) in real-world agriculture. We introduce \textbf{AgriChain}, a dataset of approximately 11,000 expert-curated leaf images spanning diverse crops and pathologies, each paired with (i) a disease label, (ii) a calibrated confidence score (High/Medium/Low), and (iii) an expert-verified chain-of-thought (CoT) rationale. Draft explanations were first generated by GPT-4o and then verified by a professional agricultural engineer using standardized descriptors (e.g., lesion color, margin, and distribution). We fine-tune Qwen-2.5-VL-3B on AgriChain, resulting in a specialized model termed \textbf{AgriChain-VL3B}, to jointly predict diseases and generate visually grounded reasoning. On a 1,000-image test set, our CoT-supervised model achieves \textbf{73.1\%} top-1 accuracy (macro-F$_1$ = \textbf{0.466}; weighted-F$_1$ = \textbf{0.655}), outperforming strong baselines including Gemini 1.5 Flash, Gemini 2.5 Pro, and GPT-4o-Mini. The generated explanations align closely with expert reasoning, consistently referencing key visual cues. These findings demonstrate that expert-verified reasoning supervision significantly enhances both accuracy and interpretability, bridging the gap between generic multimodal models and human expertise, and advancing trustworthy, globally deployable AI for sustainable agriculture. To support reproducibility and further research, the dataset and code are publicly available at \url{https://github.com/hazzanabeel12-netizen/agrichain}.
 \\ \newline \Keywords{Plant disease diagnosis, Vision–Language Models (VLMs), Chain-of-Thought reasoning.}}

\begin{document}

\maketitleabstract

\section{Introduction}

The diagnosis of fine-grained plant disease is a complex visual reasoning task: symptoms are often subtle, vary between cultivars and growth stages, and are easily confounded by abiotic stress. Although large Vision-Language models (VLMs)  demonstrate impressive general-purpose reasoning, they often lack \emph{domain-specific visual grounding} and calibrated uncertainty for agronomic imagery, leading to overconfident misdiagnoses on infected leaf images. 
This gap limits the reliability of AI-driven agricultural diagnostic—especially in arid and rapidly developing regions such as the Middle East and North Africa (MENA), where countries like the United Arab Emirates are working to strengthen domestic agriculture for food security despite water scarcity and heat stress.

Previous agricultural models such as AgriGPT-VL~\citep{yang2025agrigptvl} have advanced multimodal understanding in agriculture but focus mainly on large-scale image--text alignment rather than fine-grained diagnostic reasoning. Similarly, datasets such as PlantVillage~\citelanguageresource{PlantVillage} have enabled plant disease classification benchmarks but lack explanatory reasoning, while multimodal benchmarks such as ScienceQA~\citep{lu2022scienceqa} demonstrate that explicit chain-of-thought (CoT) supervision improves interpretability and reliability. However, no existing agricultural benchmark provides expert-authored reasoning aligned with visual evidence. As a result, current agricultural VLMs can often identify a disease but rarely explain \emph{why}, revealing a critical limitation that hinders practical adoption and undermines user trust.

To address this gap, we introduce \textbf{AgriChain}, the first dataset (to our knowledge) that pairs plant disease images with expert-verified CoT rationales and calibrated confidence labels. Our approach integrates human reasoning into model supervision: the VLM Qwen-2.5-VL-3B~\citep{bai2023qwen} is fine-tuned in approximately 11k image-rational pairs so that it learns to diagnose and justify its decisions. The rationales are refined by the agriculture engineer to ensure biological precision and an explicit link between visible symptoms (e.g., color, margin, and distribution of the lesion) and domain knowledge (e.g., viral patterns). This supervision encourages the model to focus on discriminative visual cues that mirror expert inspection.

On a held-out test set of 1,000 images spanning multiple crop–disease combinations, our reasoning-supervised model achieves \textbf{73.1\%} top-1 accuracy (vs.\ 55.8\% for the best baseline; +17.3 points), with macro-F$_1$\,=\,\textbf{0.466} and weighted F$_1$\,=\,\textbf{0.655}. The explanations generated are consistently faithful to visual evidence, improving transparency, interpretability, and auditability of decisions.

\noindent \textbf{Contributions.} This work makes three main contributions:
\begin{enumerate}\setlength{\itemsep}{0.3pt}
    \item \textbf{AgriChain Dataset:} An 11k-image dataset with expert-verified reasoning chains and calibrated confidence labels.
    \item \textbf{Reasoning-Calibrated Training:} A fine-tuning framework that jointly supervises diagnostic accuracy and rationale coherence via visual–textual alignment.
    \item \textbf{AgriReason-Bench Evaluation:} The first benchmark that assesses visual faithfulness using a Region–Text Alignment (RTA) metric, enabling quantitative interpretability analysis.
\end{enumerate}

\section{Related Work}
\label{sec:litreview}

\subsection{VLMs: Background and Evolution}
VLMs integrate visual and textual modalities to enable joint reasoning across images and natural language. Early systems (e.g., image captioning) mapped visual features to text using convolutional neural network (CNN) encoders and recurrent neural network (RNN) decoders. Transformer-based architectures fundamentally shifted the field by aligning visual and textual tokens within shared representation spaces.

CLIP~\citep{radford2021clip} demonstrated that contrastive pretraining on large-scale internet image--text pairs enables zero-shot recognition of previously unseen categories. Building on this foundation, models such as BLIP~\citep{li2022blip} and Flamingo~\citep{alayrac2022flamingo} introduced cross-attention mechanisms to better condition visual input, achieving strong multimodal performance even with limited examples. Contemporary systems such as GPT-4V~\citep{gpt4omini2024} and Gemini~\citep{gemini2024flash,gemini2024pro} can jointly process images and text, extending applications from medical imaging to agriculture.

Despite these advances, several key limitations persist: (i) \emph{domain gaps}—models trained on generic web data often underperform in specialized fields such as agriculture or medicine~\citep{bai2023domain}; (ii) \emph{hallucinations}—plausible but incorrect generations; and (iii) \emph{opacity}—a limited ability to explain predictions. These challenges highlight the need for structured reasoning to improve reliability in domain-specific contexts.

\subsection{Chain-of-Thought Reasoning in Language and Vision}
CoT prompting elicits step-by-step reasoning instead of producing only final responses. Generation of intermediate reasoning steps has been shown to improve performance in complex tasks; for example, Wei~\textit{et~al.} report substantial gains in logical, mathematical, and common sense reasoning~\citep{wei2022chain}. Benefits include greater interpretability, error traceability, and user trust; however, generated rationales may only partially reflect the true internal reasoning process of the model~\citep{Lanham2023}.

CoT methods are increasingly applied in multimodal settings. Zhang~\textit{et~al.}~\citep{Zhang2024} integrate textual and visual evidence into reasoning chains for VLMs, while Zhao~\textit{et~al.}~\citep{zhao2025planning} apply CoT prompting to embodied agents, where intermediate visual sub-goals improve planning and decision quality. Collectively, these studies demonstrate the growing potential of CoT for multimodal reasoning tasks, including fine-grained agricultural diagnostics.

\begin{figure*}[ht!]
    \centering
    \includegraphics[width=\textwidth]{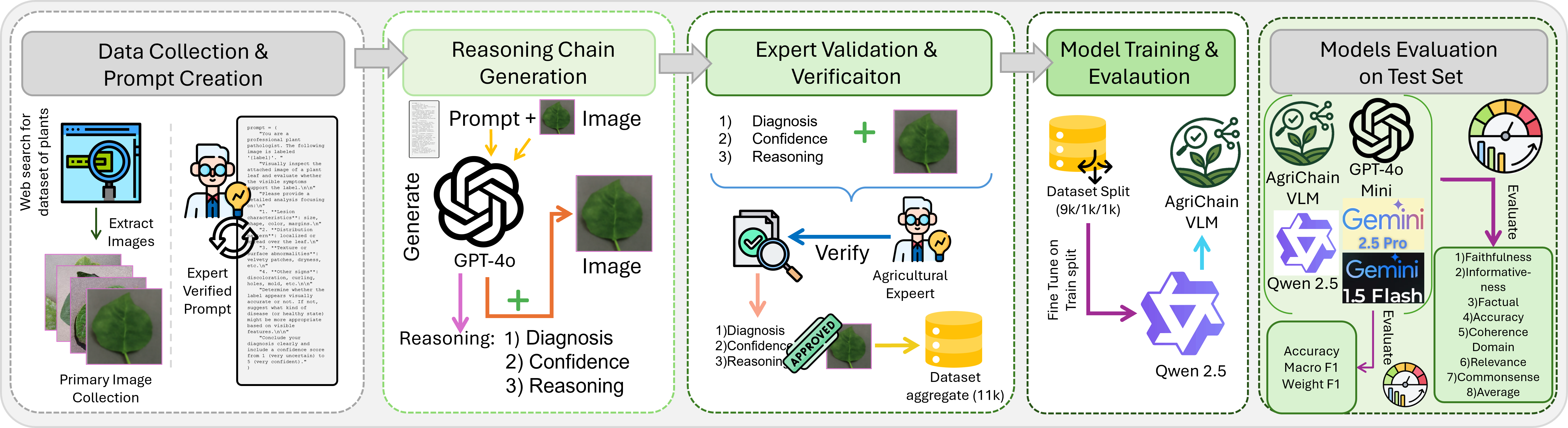}
    \caption[AgriChain training pipeline]{ AgriChain training pipeline. Images from PlantVillage~\citelanguageresource{PlantVillage}, PlantDoc \citep{singh2020plantdoc}, and PlantCLEF \citep{goeau2022plantclef} are aggregated and cleaned to form a unified dataset. A VLM (GPT-4o) drafts image-grounded diagnostic rationales, which are then verified and standardized by an agricultural expert. Finally, an open VLM (Qwen-2.5-VL-3B) is fine-tuned on these image+explanation pairs to produce both the disease prediction and an expert-style rationale for each input image.}
    \label{fig:agrichain-pipeline}
\end{figure*}

\subsection{Reasoning and Interpretability in Agricultural AI}
Although several multimodal agricultural datasets have been introduced, most focus on question--answer pairs without capturing the underlying expert reasoning process. For example, AgroInstruct ($\approx$70k QA pairs) and AgroMind ($\approx$25k QA pairs) provide valuable training data but include only answers, not the step-by-step rationales behind them. Models trained on such data can predict outcomes but cannot justify their decisions, limiting interpretability and user confidence. This gap in explainable training data motivates the development of \textbf{AgriChain}, which emphasizes expert-verified CoT explanations alongside each diagnostic prediction.

\section{Methodology}
The AgriChain methodology employs a structured multi-stage pipeline (Figure~\ref{fig:agrichain-pipeline}) that unifies expert-verified data collection, multimodal reasoning generation, and VLM fine-tuning for agricultural disease diagnosis. It begins with curated plant images and expert-designed prompts that guide detailed reasoning generation, followed by expert validation to ensure accuracy and interpretability. The verified data then support model fine-tuning and evaluation on held-out sets to assess diagnostic accuracy, confidence calibration, and reasoning faithfulness, ensuring scalability and reliability for real-world agricultural applications.

\subsection{Dataset Collection and Prompt Creation}
\subsubsection{Data Collection}
Our approach, illustrated in Figure~\ref{fig:agrichain-pipeline}, follows a structured multi-stage pipeline. We begin by collecting and curating expert-labeled plant images, after which GPT-4o is employed to generate preliminary diagnostic rationales. These explanations are subsequently reviewed and refined by agricultural experts to ensure biological precision and faithful visual grounding. The resulting expert-verified image–rationale pairs are then used to fine-tune a VLM, enabling it to produce accurate disease predictions alongside transparent, expert-level explanations.

\subsubsection{Data Aggregation and Preprocessing}
\begin{figure*}[t!]
    \centering
    \includegraphics[width=0.8\textwidth,height=9cm]{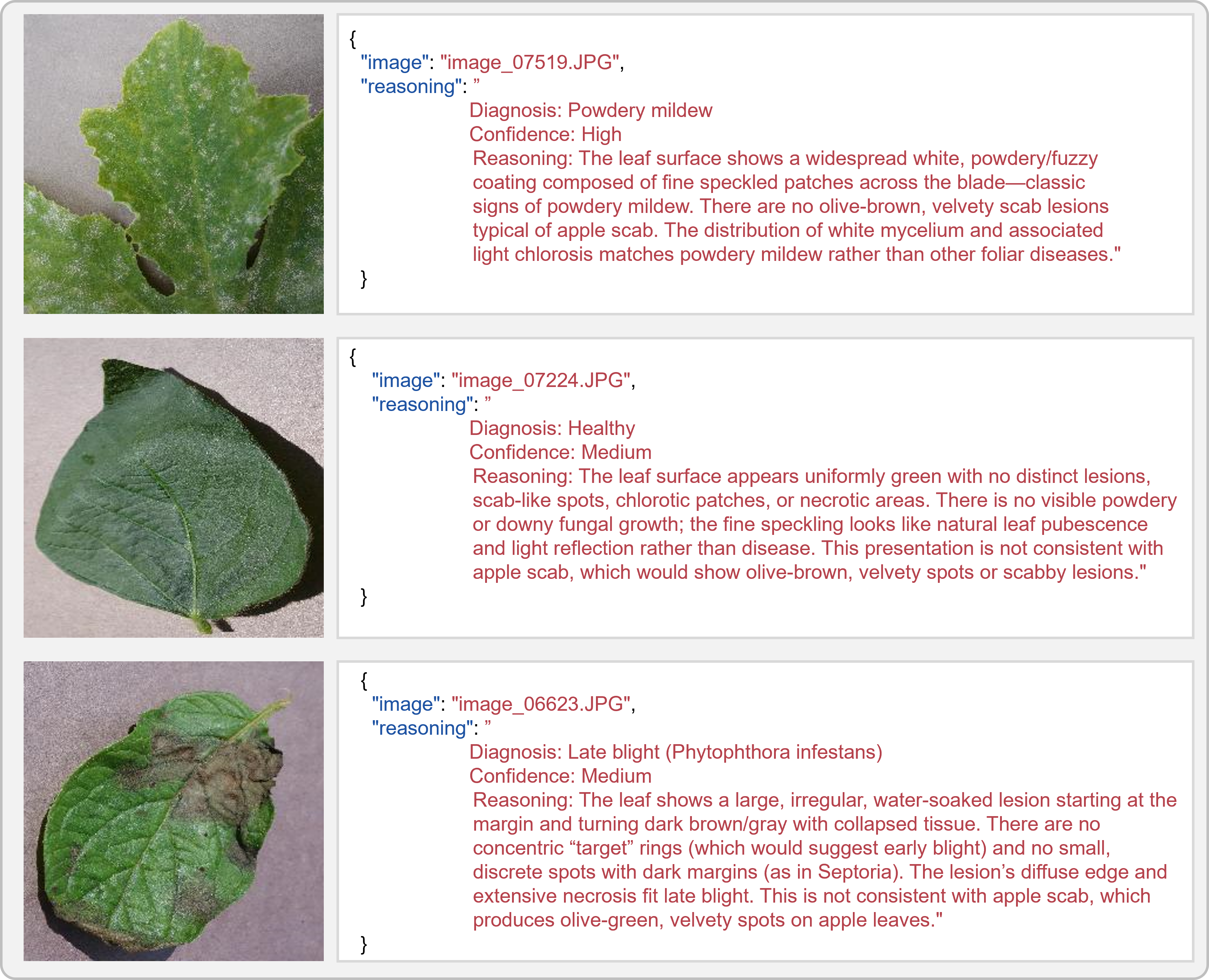}
    \caption{Example leaf disease classification outputs. Each image is paired with an expert-style diagnosis, confidence level, and reasoning generated by a vision–language model fine-tuned for agricultural disease recognition.}
    \label{fig:leaf-disease-examples}
\end{figure*}

After de-duplication, quality filtering, and harmonizing labels to a unified taxonomy, the final AgriChain dataset comprises approximately 11,000 labeled images across 33 plant-disease classes, plus a healthy class, spanning a wide range of crops (fruits such as apple, grape, citrus; vegetables such as tomato, potato, pepper; and others like banana and strawberry). Each image is annotated with its crop type and disease name, verified by expert annotation or source metadata as shown in Figure~\ref{fig:leaf-disease-examples}. All images were resized to a uniform resolution, and we applied data augmentation (random horizontal flips, small rotations, and mild color jitter) during training to improve generalization. The data was split into 9,000 training images, 1,000 validation images, and 1,000 test images, to ensure that all disease classes (and the healthy class) are represented in each split.

% \vspace{-1.0em}
\subsubsection{AgriChain Prompt}
\begin{figure}[t!]
    \centering
    \includegraphics[width=0.44\textwidth,height=12cm]{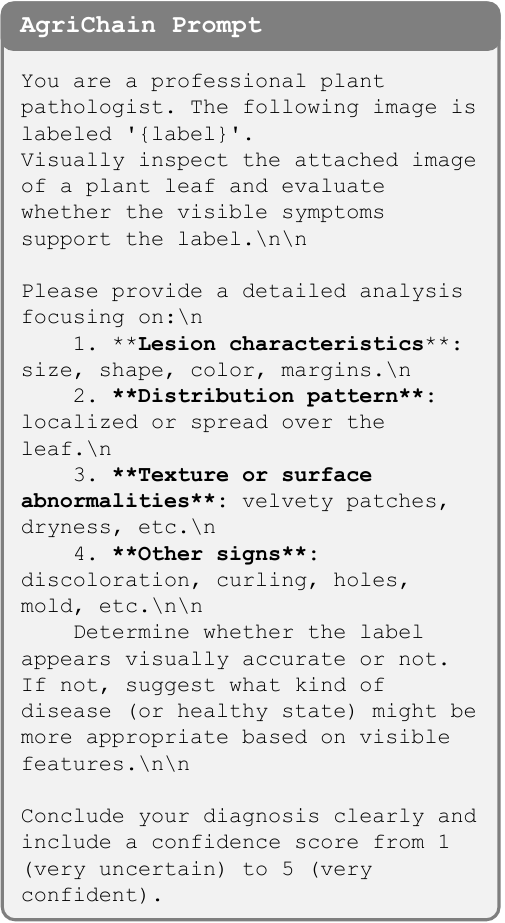}
    \caption{Prompt used in the reasoning chain generation stage, guiding the model to produce structured diagnostic analyses of plant diseases with expert-level detail and confidence scoring.}
    \label{fig:agri_prompt}
    \vspace{-0.75em}
\end{figure}

To ensure high-quality diagnostic reasoning, the prompt design of AgriChain was subjected to an iterative process of expert curation and refinement. The initial prompts were designed to mimic the analytical workflow of professional plant pathologists, guiding the model to assess the morphology, distribution, texture and visual symptoms of the lesion. Through multiple rounds of human-in-the-loop evaluation, each prompt was progressively optimized—tested, analyzed, and revised after each iteration—to enhance interpretive depth, consistency, and alignment with expert diagnostic reasoning. This iterative augmentation yielded the final version as shown in Figure~\ref{fig:agri_prompt}.

\subsection{Reasoning Chain Generation}
In the second stage of the AgriChain pipeline, reasoning chains are generated by prompting GTP-4o-mini \cite{gpt4omini2024} with curated plant images and domain-specific instructions to emulate expert diagnostic reasoning. Each response includes a structured analysis that covers the morphology, distribution, and texture of the lesion, followed by a confidence-based conclusion. These outputs are iteratively refined through human feedback to enhance clarity, faithfulness, and alignment with expert interpretations, forming the foundation for model fine-tuning and evaluation.

\vspace{-0.75em}
\subsection{Expert Validation and Verification}
For each training image, we pair the ground-truth disease label with an explanatory rationale to serve as CoT supervision. First, a strong VLM (GPT-4o) drafted a diagnostic explanation by analyzing the image’s visual cues (e.g., describing lesion color, shape, and distribution and comparing them to similar diseases). For example, given an apple leaf with scab, the model might note “orange-brown velvety lesions along the veins characteristic of apple scab”, explicitly contrasting it with absent features of other foliar diseases. 

Next, a professional agricultural engineer reviewed and refined each AI-generated rationale to ensure biological accuracy and use of standard agronomic descriptors. The expert corrected any mistakes, standardized the terminology (for instance, describing lesions with precise color and margin terms), and assigned a confidence level (High, Medium, or Low) to each diagnosis. The result is an expert-verified chain-of-thought explanation paired with each image, along with a calibrated confidence label, providing high-quality supervision for model training.

\subsection{Model Fine-Tuning}
We fine-tuned an open-source VLM, Qwen-2.5-VL-3B~\citep{Bai2023}, using the image–rationale pairs described above. The resulting model, which we refer to as \textbf{AgriChain-VL3B}, is optimized for agriculturally grounded diagnostic reasoning. To steer the model toward structured disease analysis, we adopted an \textit{instruction-following} training format in which each input combined a plant image with a concise textual prompt (Figure~\ref{fig:agri_prompt}) directing the model to identify the disease and justify its decision.

Training followed a standard autoregressive language-modeling objective, minimizing next-token prediction loss across the complete output sequence (diagnosis, confidence, and rationale). We used the AdamW optimizer with a learning rate of $2\times10\textsuperscript{-5}$, a batch size of 16, and a cosine learning-rate scheduler with 5\% warm-up steps. Early stopping based on validation loss was applied to prevent overfitting. The model exhibited stable convergence within three training epochs, after which the optimal checkpoint was retained for all subsequent evaluations. Mixed-precision (FP16) training was employed to enhance computational efficiency and minimize memory consumption without compromising model fidelity.

\section{Models Evaluation Framework}
\subsection{Zero-Shot Baseline Comparisons}
To quantify the benefit of our CoT fine-tuning, we also evaluated several strong pre-trained VLMs as zero-shot baselines under the same input conditions:
\begin{itemize}\setlength\itemsep{0em}
    \item \textbf{Gemini 1.5 Flash}: a lightweight multimodal model optimized for speed.
    \item \textbf{Gemini 2.5 Pro (Vision)}: a larger multimodal model used as a strong baseline for zero-shot diagnosis.
    \item \textbf{GPT-4o-mini}: a smaller-scale, GPT-4-style VLM.
\end{itemize}
All baseline models received the same prompt and image inputs as our fine-tuned model, but no fine-tuning was performed on the AgriChain data for these models. This allows a direct comparison of zero-shot performance versus our fine-tuned approach.

\subsection{Evaluation Criteria and Metrics}
We evaluated diagnostic performance on a held-out test set of 1,000 images, treating the task as a strict classification problem. To ensure a fair assessment across diseases, this test set was constructed to be class-balanced: it contains roughly 30 images for each of the 33 disease classes (plus the healthy class), covering a total of 34 classes. A model’s prediction for an image is counted as correct if the generated output explicitly contains the true disease name (case-insensitive match). Minor hedging or additional phrasing (e.g., “likely \textit{disease}”) is permitted as long as the correct disease name appears; however, any misidentification or failure to mention the true label is considered incorrect.

We report the following evaluation metrics for model performance:
\begin{itemize}\setlength\itemsep{0em}
    \item Accuracy – the percentage of test images for which the model’s diagnosis is correct.
    \item Macro F1 – the unweighted average of per-class F1 scores, treating each disease class equally (this emphasizes performance on minority classes).
    \item Weighted F1 - the F1 score weighted by the frequency of each class in the test set (reflecting the overall performance with the frequency of the class taken into account).
\end{itemize}
Given our uniformly distributed test set, these metrics together provide a comprehensive view of model accuracy and its robustness across both common and rare diseases.

To address this, our dataset introduces the first agricultural VQA collection with expert-verified reasoning chains, enabling evaluation of both \textbf{answer accuracy} and the \textbf{quality of the explanation}. This addition is essential for trustworthy AI in real-world agricultural contexts where decision justification matters as much as correctness.

\subsection{Benchmark Criteria for Reasoning Evaluation}
Following recent work on CoT evaluation~\citep{Lanham2023, liu2023geval}, we assess reasoning quality using six complementary criteria:
\begin{itemize}
    \item \textbf{Faithfulness}: the reasoning accurately supports the final answer.
    \item \textbf{Informativeness}: explanations provide sufficient, non-redundant detail.
    \item \textbf{Factual Accuracy}: statements are scientifically correct.
    \item \textbf{Coherence}: reasoning follows a logical and complete flow.
    \item \textbf{Domain Relevance}: knowledge used is agriculturally appropriate.
    \item \textbf{Commonsense}: explanations align with biological and logical principles.
\end{itemize}
These criteria capture both \textit{intrinsic quality} (faithfulness, informativeness, factual accuracy, coherence) and \textit{contextual appropriateness} (domain relevance, commonsense).

\subsection{Data Gathering and Expert Image Selection}

Beyond dataset aggregation, we implemented a rigorous process to ensure that only high-quality and diagnostically informative images were included. Rather than selecting samples randomly, agricultural experts systematically reviewed the candidate images from all sources (PlantVillage~\citelanguageresource{PlantVillage}, PlantDoc~\citep{singh2020plantdoc}, Roboflow Leaf Disease~\citep{roboflow2022leaf}, and PlantCLEF~2022~\citep{goeau2022plantclef}). The selection process followed three stages:

\begin{enumerate}
    \item \textbf{Preliminary Screening:} Automated scripts removed duplicates, blurred samples, and low-resolution images. 
    \item \textbf{Expert Filtering:} Two agricultural engineers independently reviewed remaining images to confirm visible, disease-specific symptoms such as lesion type, color variation, or mold growth. Images showing ambiguous or overlapping symptoms were excluded to maintain label clarity.
    \item \textbf{Final Validation:} From the filtered pool, experts curated a balanced subset per disease class, ensuring diversity across lighting conditions, growth stages, and background types. This guarantees that the dataset represents both laboratory-quality and realistic field conditions without compromising interpretability.
\end{enumerate}

This structured selection pipeline ensures that every image in the final dataset is not only correctly labeled but also visually representative of the disease, enabling models to learn discriminative and trustworthy visual cues.

\subsection{Preprocessing and Input Formatting}
After expert curation, all images were cleaned, resized to a uniform resolution, and harmonized under a single taxonomy. To support reasoning-based supervision, each instance was formatted with dual output modes:

\noindent\textbf{Input:} image $+$ natural-language prompt (e.g., ``Diagnose the disease of this plant with reasoning.'')\\
\textbf{Output:}
\begin{itemize}
    \item \textbf{Final Answer:} disease label with a confidence level.
    \item \textbf{Reasoning mode (CoT):} disease label, confidence, and a structured 3–5 step diagnostic explanation.
\end{itemize}

This dual format supports both traditional classification and reasoning-augmented training and evaluation, allowing our fine-tuned models to provide not only the correct diagnosis but also a transparent explanation grounded in visual evidence.

\section{Results, Analysis, and Insights}
Our experiments on the \textit{AgriChain} dataset show that incorporating explicit CoT reasoning significantly improves plant–disease diagnosis. Among five evaluated models, the \textbf{Agrichain-VL3B} attained the highest accuracy at \textbf{73.1\%}, outperforming \textbf{Gemini~Pro} (55.8\%), \textbf{Gemini~Flash} (48.7\%), and \textbf{GPT-4o~Mini} (34.9\%), while the baseline \textbf{Qwen-2.5-VL} performed near chance (14.4\%). AgriChain-VL3B also achieved the best F$_1$ scores (Table~\ref{tab:performance}). Although a gap remains between accuracy (73.1\%) and macro-F$_1$ (0.466), indicating tail classes are still challenging, the CoT-tuned model is consistently stronger across all metrics.

\begin{table}[th]
\centering
\setlength{\tabcolsep}{4 pt}
\caption{Comparison of model performance on the plant–disease test set. The ArgiChain-VL3B achieves the highest accuracy and F$_1$ scores, outperforming all baselines.}
\label{tab:performance}
\small
\resizebox{\columnwidth}{!}{
\begin{tabular}{l|ccc}
\toprule
\textbf{Model} & \textbf{Accuracy}$\uparrow $ & \textbf{Macro F$_1$}$\uparrow$  & \textbf{Weight F$_1$}$\uparrow$  \\
\midrule
Qwen (zero-shot)  & 14.4 & 0.023 & 0.149 \\
Gemini 1.5 Flash      & 48.7 & 0.105 & 0.430 \\
Gemini 2.5 Pro        & 55.8 & 0.141 & 0.476 \\
GPT-4o-mini      & 34.9 & 0.053 & 0.318 \\
AgriChain-VL3B\textit{(ours)} & \textbf{73.1} & \textbf{0.466 }& \textbf{0.655} \\
\bottomrule
\end{tabular}}
\end{table}

\paragraph{What the numbers mean.}
AgriChain-VL3B surpasses the next best model (Gemini~Pro) by \textbf{+17.3} accuracy points and improves macro-F$_1$ by \textbf{+0.325}, indicating better class-balanced performance. The much higher weighted F$_1$ than macro-F$_1$ (0.655 vs.\ 0.466) reveals a head–tail pattern: all models perform best on frequent diseases and struggle on rare ones, but CoT training \emph{narrows} this gap more than any baseline.

\paragraph{Why CoT helps.}
We observe three practical mechanisms by which CoT boosts performance:
\begin{enumerate}
    \item \textbf{Domain grounding.} Supervised rationales teach the model to attend to discriminative, field-relevant cues (e.g., lesion color/margin/texture, interveinal patterning), reducing shortcut features and improving label fidelity \cite{Zhang2024}.
    \item \textbf{Structured elimination.} CoT encourages explicit \emph{negative} evidence (``no yellow halos $\Rightarrow$ unlikely bacterial spot''), which helps separate visually similar diseases (e.g., downy vs.\ powdery mildew) and mitigates plausible-but-wrong guesses \cite{camburu2018}.
    \item \textbf{Better calibration and oversight.} Producing reasons and confidence promotes calibrated outputs and enables human validation; this improves trust and reduces undetected hallucinations in high-stakes use \cite{Lanham2023}.
\end{enumerate}

\paragraph{Residual errors and paths forward.}
Most remaining errors arise in (i) \textit{rare classes} with limited training support, and (ii) \textit{look-alike} foliar pathologies with overlapping symptom morphology. These failure modes suggest targeted remedies: tail-focused augmentation and sampling, contrastive/rationale-aware objectives for confusable pairs, and expanded coverage of underrepresented diseases in \textit{AgriChain}. Overall, the results substantiate that \emph{reasoning supervision}—not just labels—materially improves both accuracy and class balance, advancing practical, interpretable diagnosis in the wild \cite{Zhang2024,camburu2018,Lanham2023}.

\subsection{Explanation and Reasoning Quality}
Beyond top\mbox{-}1 accuracy, we assess \emph{how} models argue for their diagnoses. Following a rubric grounded in prior work on faithful explanations \cite{Lanham2023} and explanations-as-supervision \cite{camburu2018}, human evaluators (blinded to model identity) rated each rationale along six axes: \emph{faithfulness}, \emph{informativeness}, \emph{factual accuracy}, \emph{coherence}, \emph{domain relevance}, and \emph{commonsense} (1–5 scale; higher is better). As summarized in Table~\ref{tab:reasoning_quality}, the AgriChain-VL3B leads across nearly all criteria, averaging \textbf{4.63/5}. In contrast, the base Qwen model averages \textbf{3.40}, with short, generic rationales that lack diagnostic structure. Gemini models score strongly (Gemini~Pro: 4.55), and GPT-4o~Mini approaches AgriChain-VL3B on several dimensions, yet both remain less precise on domain-specific cues.

\paragraph{Inter-Annotator Agreement.}
To validate the reliability of our reasoning evaluation, we computed inter-annotator agreement on a randomly sampled 10\% subset of the rated rationales using Krippendorff’s $\alpha$, which is appropriate for ordinal ratings and multiple annotators. The subset was independently evaluated by three evaluators to assess annotation consistency. Agreement among annotators reached $\alpha = 0.838$ (83.8\%), indicating strong consistency and high reliability of the reasoning quality scores.

\begin{table*}[th]
\centering
\setlength{\tabcolsep}{6pt}
\caption{Average human ratings (1–5 scale) of reasoning quality across six criteria. The AgriChain-VL3B produces the most faithful, informative, and domain-relevant explanations. }
\label{tab:reasoning_quality}
\small
\resizebox{\textwidth}{!}{
\begin{tabular}{l|ccccccc}
\toprule
\textbf{Model} & \textbf{Faithfulness} & \textbf{Informativeness} & \textbf{Factual Acc} & \textbf{Coherence} & \textbf{Domain Rel} & \textbf{Commonsense} & \textbf{Average} \\
% \textbf{Model} & \textbf{Faith.} & \textbf{Info.} & \textbf{Fact.Acc.} & \textbf{Coher.} & \textbf{Dom.Rel.} & \textbf{Common.} & \textbf{Avg.} \\
\midrule
Qwen-2.5-VL(base)       & 3.5 & 3.2 & 3.4 & 3.3 & 3.0 & 4.0 & 3.40 \\
Gemini 1.5 Flash         & 4.3 & 4.2 & 4.4 & 4.3 & 4.5 & 4.5 & 4.37 \\
Gemini 2.5 Pro               & 4.5 & 4.4 & 4.6 & \textbf{4.5 }& 4.7 & 4.6 & 4.55 \\
GPT-4o Mini              & 4.5 & 4.3 & 4.6 & 4.4 & 4.6 & \textbf{4.7 }& 4.52 \\
AgriChain-VL3B\textit{(ours)} &\textbf{ 4.6} &\textbf{ 4.6} & \textbf{4.7 }& \textbf{4.5 }& \textbf{4.8} & 4.6 & \textbf{4.63} \\
\bottomrule
\end{tabular}}
\end{table*}

\paragraph{Where the gains come from.}
Qualitative analysis reveals three recurrent properties of the CoT-fine-tuned rationales that explain the score improvements:
\begin{enumerate}
    \item \textbf{Symptom anchoring and precise lexicon.} AgriChain-VL3B consistently names \emph{discriminative} visual cues (e.g., ``velvety olive\mbox{-}brown blotches'' for apple scab; ``angular chlorosis limited by veins'' for downy mildew) and links them to the diagnosis with explicit evidence–claim links. This increases \emph{faithfulness}, \emph{factual accuracy}, and \emph{domain relevance}. In contrast, base Qwen often resorts to generic phrases (``fungal infection likely'') that lack actionable detail.
    \item \textbf{Structured differential diagnosis.} The model regularly uses \emph{negative evidence} and contrasts look\mbox{-}alike conditions (e.g., ``absence of yellow halos $\Rightarrow$ unlikely bacterial spot''), which improves \emph{informativeness} and \emph{coherence} and reduces plausible-but-wrong guesses. This mirrors expert reasoning patterns emphasized in multimodal CoT studies \cite{Zhang2024}.
    \item \textbf{Calibrated, reviewable justifications.} By exposing intermediate observations and (when applicable) uncertainty, AgriChain-VL3B produces rationales that a practitioner can audit. This supports more calibrated decisions and aligns with faithfulness-focused guidance \cite{Lanham2023}.
\end{enumerate}

\begin{figure}[htbp]
\centering
\includegraphics[width=0.55\columnwidth]{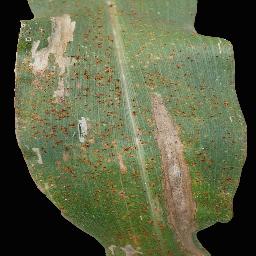}
 \caption{\textbf{Diagnosis:} Cedar-apple rust. \textbf{Confidence:} Medium. \textbf{Reasoning:} The leaf shows multiple small, circular, orange to rust-brown spots with distinct margins scattered across the blade. Lesions are not olive, velvety, or sooty, and there is no distortion or corky scabbing typical of apple scab.}
\label{fig:cedar_rust}
\end{figure}
\paragraph{Illustrative case.}
In the cedar\mbox{-}apple rust example in Fig.~\ref{fig:cedar_rust}, AgriChain-VL3B cites \emph{circular, orange to rust\mbox{-}brown lesions with clear margins} and \emph{absence of olive, velvety scab} as key evidence—precisely the cues that agronomists use to separate rust from apple scab. Baselines typically mention discoloration, but omit margin shape or vein\mbox{-}bounded patterns, reducing discriminative power.

\paragraph{Error modes and remaining gaps.}
Even with CoT supervision, AgriChain-VL3B can (i) overfit to common lexicon (e.g., overusing ``velvety''), (ii) under\mbox{-}specify atypical presentations (mixed infections, severe abiotic stress), and (iii) occasionally assert context not visible (e.g., humidity or lifecycle stage). These issues reduce \emph{faithfulness} on hard cases and suggest targeted remedies: lexicon regularization, rationale editing constraints (evidence must reference visible cues), and augmentation with rare/atypical symptom sets. Still, compared to all baselines, AgriChain-VL3B provides the most \emph{auditable} and \emph{domain-grounded} explanations, which is critical for high-stakes agricultural decisions \cite{camburu2018, Lanham2023, Zhang2024}.

\subsection{Implications for Real-World Deployment}
The superior accuracy and reasoning transparency of the CoT-enhanced model are crucial for real-world adoption. By generating explicit rationales, the model allows farmers and agronomists to understand \textit{why} a diagnosis was made—for instance, by noting that “no yellow halos are present, ruling out bacterial spot.” Such interpretability builds trust among users who might otherwise be skeptical of opaque AI systems. Including a confidence score with each prediction further enhances decision reliability, helping practitioners determine when to seek expert verification.

Moreover, model size plays an important role in deployment. Large models such as Qwen-2.5-VL and Gemini~Pro deliver the highest accuracy but demand considerable computational resources. Smaller variants like Gemini~Flash, while roughly 13 points less accurate, are more practical for on-device or offline use in remote areas. This trade-off enables flexible deployment strategies: high-capacity models can run on the cloud for large-scale analysis, while lightweight models support field-level diagnosis on mobile devices. In either case, our results emphasize that CoT reasoning improves not just accuracy but also user trust—transforming AI from a black-box classifier into a transparent decision-support tool.

\section{Limitation and Future Work}
We identify several directions for extending \textit{AgriChain} and improving reasoning-centric VLMs:
\begin{itemize}
    \item \textbf{Scalable Annotation Pipelines:} Implement semi-automated CoT labeling workflows using model-in-the-loop annotation and expert verification to accelerate dataset expansion.
    \item \textbf{Multilingual Reasoning:} Develop CoT explanations in local languages (e.g., Mandarin, Hindi, Swahili) with culturally contextual phrasing to improve accessibility and global adoption.
    \item \textbf{Expanded Coverage:} Increase the diversity of crops, diseases, and abiotic disorders (e.g., nutrient deficiencies, drought stress) to improve robustness on rare and complex conditions.
    \item \textbf{Farmer-Centric Integration:} Embed CoT-enabled diagnosis into mobile and advisory tools with offline inference and user feedback mechanisms, closing the loop between diagnosis, action, and continuous improvement.
\end{itemize}

\medskip
\noindent In summary, fine-tuning with expert-verified chain-of-thought reasoning significantly improves both accuracy and interpretability in vision–language models for agriculture. \textbf{AgriChain} represents a step toward transparent, trustworthy, and field-ready AI systems that empower human decision-makers while ensuring scientific reliability.
\section{Conclusion}
\label{sec:conclusion}

\noindent

\noindent
We introduced \textbf{AgriChain}, an expert-curated dataset that pairs plant-disease images with \emph{chain-of-thought} rationales and calibrated confidence labels. Training a VLM on AgriChain resulted in a specialized model, \textbf{AgriChain-VL3B}, which significantly improved both accuracy and interpretability, achieving state-of-the-art results (73.1\% accuracy; macro-F$_1$\,=\,0.676; weighted-F$_1$\,=\,0.655), surpassing strong baselines such as Gemini Pro (55.8\%). The explanations generated by AgriChain-VL3B consistently referenced key visual cues (e.g., lesion color, margin, and distribution), confirming that reasoning supervision helps models \emph{think like experts} and produce more reliable diagnoses.

\noindent
Beyond metrics, interpretability is vital for real-world use. Farmers and agronomists must understand \emph{why} a diagnosis is made. CoT supervision addresses this by revealing the same visual evidence experts rely on, turning the model into a transparent decision-support partner rather than a black box. The approach is broadly applicable—domain-specific reasoning can transform general VLMs into trustworthy specialists for related tasks such as pest or nutrient deficiency detection. Together, AgriChain and CoT fine-tuning pave the way for field-ready AI systems that are accurate, transparent, and easy to trust.

\section{Bibliographical References}\label{sec:reference}

\bibliographystyle{lrec2026-natbib}
\bibliography{lrec2026-example}

\label{lr:ref}
\bibliographystylelanguageresource{lrec2026-natbib}
\bibliographylanguageresource{languageresource}

\end{document}